\documentclass{article}
\usepackage{spconf,amsmath,graphicx}
\usepackage{booktabs}
\usepackage{amsmath}
\usepackage{hyperref}
\usepackage{amssymb}
\usepackage{enumitem}
\usepackage{xcolor}
\usepackage{color, colortbl}
\usepackage{tabularx}
\usepackage{caption}

\title{L1-aware Multilingual Mispronunciation Detection Framework}

\name{Yassine El Kheir, Shammur Absar Chowdhury, Ahmed Ali}
\address{Qatar Computing Research Institute, HBKU, Doha, Qatar}
\begin{document}
%
\maketitle
\begin{abstract}
The phonological discrepancies between a speaker's native (L1) and the non-native language (L2) serves as a major factor for mispronunciation. 
This paper introduces a novel multilingual MDD architecture, \textbf{L1-MultiMDD}, enriched with L1-aware speech representation. 
An end-to-end speech encoder is trained on the input signal and its corresponding reference phoneme sequence. First, an attention mechanism is deployed to align the input audio with the reference phoneme sequence. Afterwards, the L1-L2-speech embedding are extracted from an auxiliary model, pretrained in a multi-task setup identifying L1 and L2 language, and are infused with the primary network. Finally, the 
L1-MultiMDD is then optimized for a unified multilingual phoneme recognition task using connectionist temporal classification (CTC) loss for the target languages: English, Arabic, and Mandarin.  
Our experiments demonstrate the effectiveness of the proposed L1-MultiMDD framework on both seen -- L2-ARTIC, LATIC, and AraVoiceL2v2; and unseen -- EpaDB and Speechocean762 datasets. 
The consistent gains in PER, and false rejection rate (FRR) across all target languages confirm our approach's robustness, efficacy, and generalizability.

\end{abstract}
\section{Introduction}
\label{sec:intro}
Mastering the second language (L2) pronunciation is crucial for effective communication and cultural integration. Mispronunciation detection and diagnosis (MDD) systems are integral parts of the Computer-aided Pronunciation Training (CAPT) system. Its effectiveness in aiding non-native speakers to overcome the challenges of learning a new language, and improve pronunciation skills \cite{article}. 

Various approaches have been investigated to automate the MDD system. Methods extended from relying on pre-trained automatic speech recognition (ASR) systems to identify discrepancies between the aligned ASR output and the reference sequence, utilizing the log-posterior probability from the ASR to calculate different measures of goodness of pronunciation (GOP) scores \cite{GOP_WITT, GOP_1, GOP_2} are explored. 
The deep learning techniques, either in an end-to-end approach or by using a cascaded pipeline -- trained with GOP features from a pre-trained ASR \cite{JIM,3M}, have also been utilized. The end-to-end models trained using the Connectionist Temporal Classification (CTC) loss have become increasingly popular due to their impressive performance in tasks related to detecting mispronunciations without the need for any alignment information \cite{non_ASR, feng2020sed, wu2021transformer, fu2021full, xu2021explore, kheir2023multi}. 

With recent advancement in multilingual speech research \cite{pratap2023scaling}, there have been numerous efforts to model pronunciation assessment on utterance level scores in a multilingual context. Studies such as \cite{zhang2021multilingual} explores the adaptation of pronunciation assessments from English (a stress-timed language) to Malay (a syllable-timed language), and \cite{lin2023multi} investigates the use of language-specific embeddings for diverse languages, while optimizing the entire network within a unified framework. 
However, limited research investigated MDD in multilingual scenarios.






In this paper, we introduce a novel L1-aware multilingual MDD -- \textbf{L1-MultiMDD} -- framework. The proposed framework exploits the contrasting and shared phonological representation of various L2 languages while considering the speaker's L1 background. The framework is comprised of a primary MDD network and an auxiliary network designed to encode L1-L2 aware representation. The auxiliary network is instrumental in extracting and fusing L1-L2 language-specific information in the multilingual latent space, allowing the L1-MultiMDD to become more sensitive to the idiosyncrasies of each respective language.

The proposed L1-MultiMDD model supports $3$ target languages -- English, Arabic, and Mandarin, and is designed to accommodate speakers from any L1 background. The proposed model is superior in performance to monolingual models, with better generalization.
Compared to prior research, our study is the first to investigate the role of multilinguality in MDD utilizing speakers' L1 background. This integration of L1-L2 representation not only enriches the MDD performance but also opens up new avenues for \textit{personalized and effective pronunciation assessment techniques}.

\begin{figure*} [!ht]
\centering
\vspace{-0.4cm}
\includegraphics[width=0.8\textwidth]{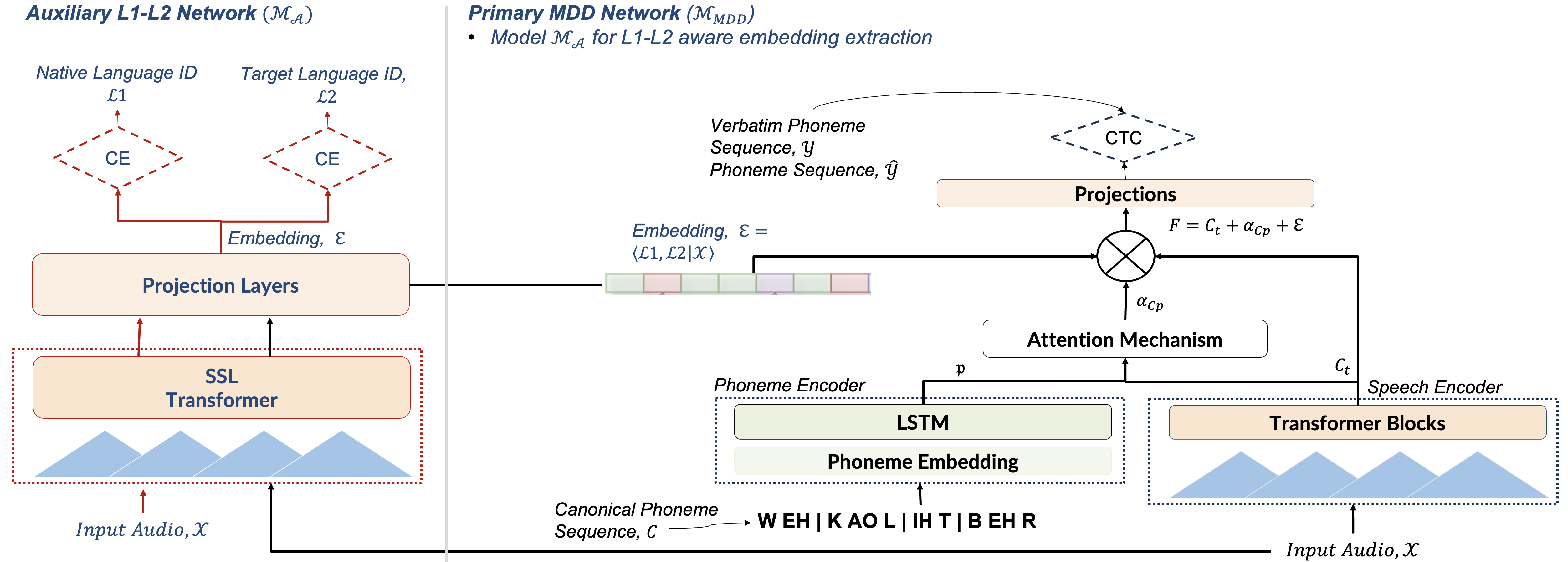}
\vspace{-0.2cm}
\caption{\small{Overview of proposed L1-MultiMDD framework.}}
\label{fig:model}
\vspace{-0.2cm}
\end{figure*}

\begin{table*}[!ht]
\centering
\scalebox{0.7}{
\begin{tabular}{lccccc}
\toprule
\multicolumn{1}{c}{\textbf{Corpus}} & \textbf{Target (L2)} & \textbf{Native (L1)} & \textbf{\#Spks} & \textbf{Use (Unseen)} & \textbf{Split(trn/vld/tst)} \\ \midrule\midrule
L2-ARTIC \cite{L2-ARCTIC} & English & Hindi, Korean, Mandarin, Spanish, and Arabic & 24 & trn/tst & $12(2hrs)/6(0.8hrs)/6(0.8hrs)$ \\
LATIC \cite{mqtj-qh10-21}& Mandarin & Russian, Korean, French,and Arabic & 4 & trn/tst & $2(2hrs)/1(0.8hrs)/1(1hr)$ \\
\begin{tabular}[c]{@{}l@{}}AraVoiceL2 \cite{speechblender} \\      \space v2*\end{tabular} & Arabic & \begin{tabular}[c]{@{}c@{}}Turkish, Benin, Nigerian, Ivory, Indonesian, Thailand, \\ Malaysian, Bangladeshi, Madagascar, Chad, and Cameroon\end{tabular} & 32 & trn/tst & $19(9hrs)/2(0.8hrs)/11(5.15hrs)$\\ \midrule
EpaDB \cite{epadb}& English & Spanish & 50 & tst (Domain) & $1263$ \#utt\\
Speechocean762 \cite{speechocean762}& English & Mandarin & 250 & tst (C and A) & $2500$ \#utt\\
\bottomrule
\end{tabular}}

\caption{\small{Data description for L1-MultiMDD model.
\#Spks: Number of speakers.\#utt: Number of utterances. C: Children and A: Adult data. 
}}
\vspace{-0.3cm}
\label{tab:dataset_tab}
\end{table*}

\section{Proposed L1-MultiMDD Framework}
\label{sec:method}

Figure \ref{fig:model} shows an overview of our framework, designed to train an E2E-MDD model using multiple sources of information. The system is comprised of a primary MDD network and an auxiliary network. 

The \textit{primary MDD} model adopts the phoneme recognition task as the main objective, and enforces that the network is able to detect the verbatim phoneme sequence of the speaker's utterance. The primary MDD network takes raw input speech, along with canonical phonemes of the reference text. 
The \textit{auxiliary network} encapsulates the speaker-invariant representation by classifying L1 and L2 languages of the speaker in a supervised manner. The network is text-independent and takes raw speech as input to the network.

\subsection{Primary MDD Network}
Given the input raw signal $\mathcal{X} = (x_1, x_2, ..., x_n)$, of \textit{n} sample length, we first extracted contextualized acoustic representations $C_t$ (of dimension $D: 1024$) from the \textit{Speech Encoder}.
The speech encoder uses a CNN-based encoder network to encode the raw audio sample and a transformer-based context network to build context representations $C_t$ over the entire latent speech representation $Z$. 
The weight of the speech encoder is initiated from the pre-trained network trained in self-supervised manner with a weighted sum of contrastive loss and code-book diversity loss.
Simultaneously, we extract the canonical phoneme embedding $\mathfrak{p}$ from the \textit{Phoneme Encoder} which is composed of an embedding layer ($D:82$)\footnote{For monolingual setups, we used the dimension of language-specific phoneme set} followed by a BiLSTM layer ($D: 512$) to give $\mathfrak{p}$ ($D: 1024$).

The phoneme embedding $\mathfrak{p}$ and the contextualized acoustic representations $C_t$ are then passed to an attention layer that takes $\mathfrak{p}$ as key and query, and $C_t$ as value.
The embeddings $C_t$ and $\alpha_{Cp}$ ($D: 1024$) are then concatenated with the L1-L2-aware embedding $\varepsilon$ ($D: 1024$) from the auxiliary network to make the L1-L2-aware latent representation, $F = C_t + \alpha_{Cp} + \varepsilon$ ($D: 3072$). The $F$ is passed through a feed-forward projection layer and then to a phoneme output head ($PR$). The output head is trained using the CTC loss ($\mathcal{L}_{PR}$), calculated by comparing the predicted L2 speech phoneme sequence $\hat{\mathcal{Y}} = (\hat{y}_1, \hat{y}_2, ..., \hat{y}_m)$ with the human-labeled phoneme sequences $\mathcal{Y} = (y_1, y_2, ..., y_l)$.

\paragraph*{Multilingual Phoneme Set}
We construct a unified phoneme set combining all the phoneme units present in the three target languages. We prepared a list of phonemes $\mathcal{PS}_l$ that are present in the canonical and the verbatim phoneme sequences of the target languages, $l \in \{\textbf{E}nglish, \textbf{M}andarin, \textbf{A}rabic\}$. Hence the final list is $\mathcal{PS} = \mathcal{PS}_{E} \cup \mathcal{PS}_{M} \cup \mathcal{PS}_{A}$, including all the common phoneme symbols while preserving language-specific phonemes. We used $\mathcal{PS}$ ($D: 82$) in both input and output level for the MDD model. We obtain the phonemic annotations of reference and manually annotated transcriptions using ESpeak\footnote{https://github.com/espeak-ng/espeak-ng}.

\subsection{Auxiliary L1-L2 Network}
The auxiliary network takes raw signal $\mathcal{X}$ as input and passes it to a pre-trained SSL model, containing a CNN-based feature encoder and transformer encoders, to extract the contextualized acoustic features. These features are then passed through a projection layer to get L1-L2 aware \textbf{auxiliary embedding} $\varepsilon$, followed by task-dependent output head. The model $\mathcal{M_A}$ is optimized for \textit{Native} and \textit{Target} language identification jointly using Cross-Entropy (CE) as loss functions. 

\noindent \textbf{One-hot representation of L1-L2} ($\mathbb{O}$): We also examined the impact of incorporating L1-L2 information through one-hot encoding instead of auxiliary embedding. This implementation is straightforward and is most effective when all L1 background are known.



\subsection{Parameter Learning Strategies}

During parameter learning, both the networks, $\mathcal{M_{MDD}}$ and $\mathcal{M_A}$ are either jointly or sequentially optimized. For \textbf{Sequential} learning, the auxiliary $\mathcal{M_A}$ network is first trained independently. We then freeze the auxiliary model and use it to extract L1-L2 aware embedding ($\varepsilon$) for the primary $\mathcal{M_{MDD}}$ network training.
For the \textbf{Joint} modeling, we initiate the $\mathcal{M_{A}}$ using the pre-trained weights, we then update the weights of both the $\mathcal{M_{MDD}}$ and $\mathcal{M_{A}}$ models during training.  $\varepsilon^*$ represents the final embedding from $\mathcal{M_{A}}$, the jointly learned with $\mathcal{M_{MDD}}$ model.

\section{Experimental Setup}
\label{sec:exp}

\subsection{Datasets}
\label{sec:datasets}
To train the multilingual model, we combined the monolingual MDD dataset, presented in Table \ref{tab:dataset_tab}, and created multi-target MDD dataset. For the auxiliary $\mathcal{M_A}$ model, we deliberately avoided using any additional L2 data and used the same dataset as the primary MDD model, Nonetheless, our L1-MultiMDD model remains flexible, readily accommodating to any external data or pre-trained models.

\subsection{Pre-trained Speech Encoder}
The wav2vec2-large\footnote{https://huggingface.co/facebook/wav2vec2-large-robust} \cite{conneau2020unsupervised} model is a pre-trained wav2vec2.0
It follows the same architecture as the wav2vec2.0 model. The encoder network consists of blocks of temporal convolution layers with $512$ channels, and the convolutions in each block have strides and kernel sizes that compress about $25$ms of $16$kHz audio every $20$ms. The context network consists of $24$ blocks with model dimension $1024$, inner dimension $4096$, and $16$ attention heads. 

\subsection{Model Training}
\label{sec:model_training}
The models are optimized using Adam optimiser for 100 epochs with early stopping criterion. The initial learning rate is set to $1 \times 10^{-4}$, with a batch size of $32$. For all setups, the features extraction layers remain frozen and only transformer layers are finetuned. The primary loss criterion is CTC loss denoted as $\mathcal{L}_{PR}$ with a weight $\alpha =1$. However, in the case of L1-Multi:$<\mathfrak{p},\varepsilon^*>$ (Joint) setup, a combination loss is utilised consisting of $\mathcal{L}_{PR}$ with weight $\alpha =0.8 $ and CE losses for classification of L1 and L2 denoted as $loss_{L1}$, $loss_{L2}$ with a weight $1 -\alpha =0.2$. The total loss is then defined then as $\mathcal{L}_{total}$ = $\alpha \times \mathcal{L}_{PR} + (1-\alpha) \times (loss_{L1} + loss_{L2})$. Using the same configurations, we conducted fine-tuning on the auxiliary L1-L2 Network $\mathcal{M_A}$, which involved optimizing it with a weighted loss. This weighted loss was defined as $(0.5 \times loss_{L1} + 0.5 \times loss_{L2})$. 
$\mathcal{M_A}$ achieves L1 classification accuracy of 60\%, and L2 classification accuracy of 99.9\% evaluated on common testset described in Section \ref{sec:datasets}. 



\begin{table}[]
\centering
\scalebox{1.}{
\begin{tabular}{lccc}
\toprule
Dataset & F1 $\uparrow$ & FRR $\downarrow$ & PER $\downarrow$\\ \toprule
\midrule
\rowcolor{blue!10}
\multicolumn{4}{c}{Monolingual {\em vs} Multilingual} \\\midrule
\midrule
\multicolumn{4}{c}{Mono} \\
\midrule

L2-Artic & 56.42\% & 6.30\% & 14.13 \\
LATIC & 56.13\% & 7.25\% & 9.27 \\
AraVoiceL2 v2 & 76.91\% & 3.95\% & 7.39 \\
\midrule

\multicolumn{4}{c}{Mono:$<\mathfrak{p}>$} \\ \midrule
L2-Artic & \underline{57.37}\% & 5.61\% & 13.46 \\
LATIC & 73.91\% & 3.11\% & 4.61 \\
AraVoiceL2v2 & \underline{77.73}\% & 2.93\% & 7.03 \\

\midrule
\multicolumn{4}{c}{Multi} \\
\midrule

L2-Artic & 54.40\% & 5.76\% & 13.99 \\
LATIC & 54.40\% & 7.82\% & 10.16 \\
AraVoiceL2v2 & 75.86\% & 3.74\% & 7.42 \\
\midrule
\multicolumn{4}{c}{Multi:$<\mathfrak{p}>$} \\ \midrule
L2-Artic & 55.81\% & 5.46\% & 13.70 \\
LATIC & 66.84\% & 4.25\% & 6.36 \\
AraVoiceL2v2 & 71.76\% & 4.62\% & 10.96 \\ \midrule
\multicolumn{4}{c}{Multi:$<\mathfrak{p}, \mathbb{O}(L2)>$} \\ \midrule
L2-Artic & 55.98\% & \underline{5.10}\% & \underline{13.29} \\
LATIC & \underline{75.65}\% & \underline{2.54}\% & \underline{4.11} \\
AraVoiceL2v2 & 76.88\% & \underline{2.46}\% & \underline{6.86} \\
\midrule\midrule
\rowcolor{green!10}
\multicolumn{4}{c}{Proposed L1-aware Multi:One-hot Embed. (topline)} \\
\midrule
\midrule
\multicolumn{4}{c}{L1-Multi:$<\mathfrak{p},\mathbb{O}(L1)>$} \\\midrule
L2-Artic & \textit{57.14}\% & 4.56\% & 12.66 \\
LATIC & \textbf{75.96}\% & 2.32\% & \textbf{3.92} \\
AraVoiceL2v2 & 76.28\% & 2.31\% & 6.80 \\\midrule
\multicolumn{4}{c}{L1-Multi:$<\mathfrak{p},\mathbb{O}(L1,L2)>$} \\\midrule
L2-Artic & 57.12\% & \textbf{4.26}\% & \textbf{12.52} \\
LATIC & 75.77\% & \textit{2.27}\% & 3.93 \\
AraVoiceL2v2 & \textit{78.00}\% & \textbf{2.03}\% & \textit{6.42} \\\midrule
\midrule
\rowcolor{green!10}
\multicolumn{4}{c}{Proposed L1-aware Multi:Aux Embed.} \\\midrule
\midrule
\multicolumn{4}{c}{L1-Multi:$<\mathfrak{p},\varepsilon>$ (Sequential)} \\\midrule
\rowcolor{yellow!20}
L2-Artic & \textbf{57.40}\% & 4.60\% & 12.55 \\
\rowcolor{yellow!20}
LATIC & 74.82\% & 2.45\% & 4.19 \\
\rowcolor{yellow!20}
AraVoiceL2v2 & \textbf{78.42}\% & 2.25\% & \textbf{6.13 }\\\midrule
\multicolumn{4}{c}{L1-Multi:$<\mathfrak{p},\varepsilon^*>$ (Joint)} \\\midrule
L2-Artic & 56.87\% & 4.87\% & 12.92 \\
LATIC & \textit{75.34}\% & \textbf{2.23}\% & 4.38 \\
AraVoiceL2v2 & 76.44\% & 2.27\% & 6.35 \\
\bottomrule
\end{tabular}}
\vspace{-0.2cm}
\caption{\small{Results for different experimental settings. $<\mathfrak{p}>$ represent phoneme embedding, $<\mathbb{O}(L_*)>$: indicate one-hot $L_*$. Auxiliary (Aux.) network training sequentially ($<\varepsilon>$) or jointly ($<\varepsilon^*>$). topline: best-case scenario knowing all possible L1s. Green rows highlight our proposed frameworks; bold values are best for each dataset/metric.
}}
\vspace{-1.6cm}
\label{tab:seen_rslt}
\end{table}

\subsection{Evaluation}
We assess the model's ability to predict L2 phoneme sequences by reporting phoneme error rate. 
Additionally, we adopt the proposed hierarchical evaluation structure outlined in \cite{evaluation_metrics} to report F1 score, which takes into account the predicted, reference and annotated phoneme sequences, and it's computed using precision indicating the number of modeled mispronunciations over the number of mispronunciations returned by the rule set, and the recall representing the number of modeled mispronunciations over
the number of mispronunciations found in the evaluation set. Furthermore, we reported false rejection rate (FRR), which signifies the rate of phonemes recognized as mispronunciations when the actual pronunciations are correct. FRR measure the model's ability to distinguish between close phoneme among different languages in our multilingual setup.
Hence FRR and PER are our primary evaluation metrics.


\section{Results and Discussion}
\label{sec:result}

Table \ref{tab:seen_rslt} presents the efficacy of our L1-MultiMDD model over both monolingual and multilingual systems with phoneme embedding.

\noindent\textbf{Monolingual {\em vs} Multilingual MDD systems} Our results indicate that monolingual with (or without) phoneme embedding ($\mathfrak{p}$) outperform multilingual models lacking target language (L2) information. The multilingual model struggles to distinguish between the close sound units without any L2 knowledge. This highlights the importance of providing L2 information to the network, this observation is also supported in Multi:$<\mathfrak{p}, \mathbb{O}(L2)>$ results where addition of one-hot encoded L2 information in the multilingual system drastically improve PER and FFR. Showing $\Rightarrow $
\textit{Multi:$(\mathfrak{p}, \mathbb{O}(L2))$ $\gg$\footnote{$\gg$ represent the strenght of the architecture.} Mono:$(*)$}

\noindent\textbf{Efficacy of L1 and L2 information induction} When comparing between the Multi:$<\mathfrak{p}, \mathbb{O}(L2)>$ model with its L1-aware counterpart L1-Multi:$<\mathfrak{p}, \mathbb{O}(L1,L2)>$, we observed a significant improvement in all three test datasets. This trend remains consistant when comparing 
Multi:$<\mathfrak{p}, \mathbb{O}(L2)>$ with L1-aware model using auxiliary network (L1-Multi:$<\mathfrak{p},\varepsilon>$). These results highlights the significance of incorporating L1-awareness into MDD models, leading to conclusion $\Rightarrow$ L1-Multi:$(*)$ $\gg$ $\{$Multi:$(*)$, Mono:$(*)\}$



\noindent\textbf{One-hot {\em vs} Auxiliary embedding} We investigate the effects of incorporating L1-L2 information into MDD model through two methods: one-hot embedding ($\mathbb{O}$) and embedding ($\varepsilon$) extraction from the auxiliary model. Our finidings indicate that $\mathbb{O}$ performs slightly better than Auxiliary embedding, $\varepsilon$ in terms of performance. However, using $\mathbb{O}$ comes with a cost of limiting L1-MultiMDD model to few L1 backgrounds. Hence, unless the focused learner has specific L1 background, auxiliary embedding shows superior performance. 

\begin{table}[!ht]
\scalebox{0.75}{
\begin{tabular}{lcp{1.5cm}||cc}
\toprule
\multicolumn{1}{c}{\textit{Unseen dataset}} & \multicolumn{2}{c}{\textit{Speechocean762}} & \multicolumn{2}{c}{\textit{EpaDB}} \\\hline
\multicolumn{1}{c}{\textbf{Models}} & FRR $\downarrow$ & PER $\downarrow$ & FRR $\downarrow$ & PER $\downarrow$\\ \midrule\midrule
Mono:$<\mathfrak{p}>$[L2-ARTIC] & 21.30\% & 28.9 & 13.32\% & 27.56 \\
Multi:$<\mathfrak{p},\mathbb{O}(L1)>$ & 18.65\% & 22.95 & 13.01\% & 26.35 \\
\rowcolor{yellow!20}
L1-Multi:$<\mathfrak{p},\mathbb{O}(L1,L2)>$ & 15.41\% & 19.6 & 10.12\% & 23.82 \\
\rowcolor{yellow!20}
L1-Multi:$<\mathfrak{p},\varepsilon>$ & 15.56\% & 19.62 & 10.93\% & 24.82 \\
L1-Multi:$<\mathfrak{p},\varepsilon^*>$ & 19.27\% & 23.82\% & 14.54\% & 27.82\%\\
\midrule
Upper Bound Mono & -- & 9.93\%$^+$\cite{ryu23_interspeech} & -- & 17.67\%$^*$ \\

\bottomrule
\end{tabular}}
\vspace{-0.2cm}
\caption{\small{Reported FRR and PER for unseen dataset. Mono model: trained using L2-ARTIC data set. $\#^+$: The model is jointly learned for MDD and pronunciation assessment in a multi-task setting with additional L1 data. $\#^*$: Finetuned wav2vec2-large model on EpaDB.}}
\vspace{-0.2cm}
\label{tab:unseen_rslt}
\end{table}

\noindent\textbf{Learning L1-L2 jointly {\em vs} sequentially} As for the training strategy of the auxiliary network, our results shows sequential learning to be more effective than joint training in terms of performance when both trained on a constrained setup. 

\noindent\textbf{Generalization of the Framework} Table \ref{tab:unseen_rslt} shows FRR and PER results for two unseen datasets: SpeechOcean762 and EpaDB, both featuring English as L2. EpaDB has an unseen L1 of Argentinian Spanish, while SpeechOcean762 has L1 Mandarin and includes both adult and child speakers—data types our multilingual MDD models haven't previously encountered. Despite these challenges, we observed that L1-MultiMDD models significantly outperform monolingual English model trained with L2-ARTIC, indication the generalization potential of the proposed network.

\section{Conclusion}
In this study, we propose a novel L1-aware multilingual MDD, L1-MultiMDD, model that (a) leverage an auxiliary network to capture speakers L1 background jointly with the L2 language information; (b) fuse the extracted L1-L2 aware embedding to the MDD model; and (c) learn to capture multilingual latent representation while being sensitive to uniqueness of each language. Our proposed L1-MultiMDD demonstrate efficacy over its counterpart MultiMDD, and outperforms all variation of monolingual models, trained separately on the in-domain data. The L1-MultiMDD shows better generalization capability when compared to monolingual model trained on unseen data. 

\bibliographystyle{IEEEbib}
\bibliography{qvoice}

\end{document}